\documentclass[conference]{IEEEtran}
\IEEEoverridecommandlockouts
\usepackage{cite}
\usepackage{amsmath,amssymb,amsfonts}
\usepackage{graphicx}
\usepackage{textcomp}
\usepackage{multirow}
\usepackage{booktabs}
\usepackage{epstopdf}
\usepackage{algorithm}
\usepackage[algo2e]{algorithm2e} 
\usepackage{algpseudocode}
\usepackage{url}
\usepackage{hyperref}
\usepackage[utf8]{inputenc}
\usepackage[english]{babel}
\usepackage[belowskip=-4pt,aboveskip=0pt]{caption}
\begin{document}
\title{\Large{\textbf{Diving Deep onto Discriminative Ensemble of Histological Hashing \& Class-Specific Manifold Learning for Multi-class Breast Carcinoma Taxonomy*}}}
\author{\IEEEauthorblockN{Sawon Pratiher$^\S$ and Subhankar Chattoraj$^\ddagger$}
\IEEEauthorblockA{$^\S$\textit{Department of Electrical Engineering, Indian Institute of Technology Kharagpur, WB, India}\\
$^\ddagger$\textit{Department of Electronics \& Communication Engineering, Techno India University, WB, India}\\
\thanks{*This paper is accepted for presentation at 44$^{th}$ International Conference on Acoustics, Speech, \& Signal Processing (IEEE ICASSP), UK, 2019}
}}
%%%%%%%%%%%%%%%%%%%%%%%%%%%%%%%
\maketitle
\begin{abstract}
Histopathological images (HI) encrypt resolution dependent heterogeneous textures \& diverse color distribution variability, manifesting in micro-structural surface tissue convolutions \& inherently high coherency of cancerous cells posing significant challenges to breast cancer (BC) multi-classification. As such, multi-class stratification is sparsely explored \& prior work mainly focus on benign \& malignant tissue characterization only, which forestalls further quantitative analysis of subordinate classes like adenosis, mucinous carcinoma \& fibroadenoma etc, for diagnostic competence. In this work, a fully-automated, near-real-time \& computationally inexpensive robust multi-classification deep framework from HI is presented. 

The proposed scheme employs deep neural network (DNN) aided discriminative ensemble of holistic class-specific manifold learning (CSML) for underlying HI sub-space embedding \& HI hashing based local shallow signatures. The model achieves 95.8\% accuracy pertinent to multi-classification, an 2.8\% overall performance improvement \& \textbf{38.2\%} enhancement for Lobular carcinoma (LC) sub-class recognition rate as compared to the existing state-of-the-art on well known \textit{BreakHis} dataset is achieved. Also, 99.3\% recognition rate at 200$\times$ \& a sensitivity of 100\% for binary grading at all magnification validates its suitability for clinical deployment in hand-held smart devices.
\end{abstract}
\vspace{0.2cm}
%%%%%%%%%%%%%%%%%%%%%%%%%%%%%%%%%%%%%%%%%%%%%%
\section{INTRODUCTION}
Burgeoning cancer statistics in 2018 from World Cancer Report (WCR) estimate 2 million new cases of BC being registered worldwide [1]. World Health Organization (WHO) projects 627,000 women died from BC (which is approximately 15\% of all cancer related deaths among women in 2018) \& impacts 2.1 million women each year [2]. Significant research spanning imaging technique like computed tomography (CT), mammography, HI anatomization \& magnetic resonance (MR) have been developed for early-stage BC prognosis through precision medicine initiative, but HI analysis is taken as the "gold standard" due to its rich encoded histological morphology \& substantiating abnormal cellular activity. Additionally, deformation dynamics of these spatial tissue textures allows more specific characterizations from a diagnostic perspective [3] \& aid pathologists to control growth \& metastasis of tumor cells \& devise therapeutic clinical schedules which are specific to particular sub-classes.

However, owing to ubiquitously abundant inhomogeneous morphology \& complex intricate spatial correlations in the underlying inter-weaved biological tissue fabric of biopsy samples [4], automated machine vision for robust \& accurate multi-class BC detection is still a challenging task \& eludes researchers. Manual BC multi-classification by the pathologist is arduous \& requires domain expertise with interpretations being subjective in nature. An automated computer-aided diagnostic (CAD) system overcome these challenges \& assist clinicians in reliable diagnosis by reducing their workload \& avoid erroneous diagnosis [5-6]. Further, practical CAD systems face challenges of inefficient hand-crafted feature engineering, which is computationally intensive \& time consuming. Secondly, optimal supervised feature selection for accurate BC identification via image segmentation \& identification of primitives like nuclei deformation, tubule formation, lymphocytes presence etc.,[7] \& high-resolution HI analysis is computationally expensive requiring costly high-performance computation, which is generally not available in developing countries. Subtle inter-class \& intra-class variability w.r.t., contrast \& textures are evident in fig. 1. Fig. 1 shows representative Hematoxylin \& Eosin (H\&E) stained fine-grained multi-class biopsy tissue slides. (Taken from \textit{BreaKHis} @$400\times$ magnification [8]).  
%%%%%%%%%%%%%%%%%%%%%%%%%%%%%%%%%%%%%%%%%%%%%%
\begin{figure}[!ht]
\centering
\includegraphics[width=9cm, height=4cm]{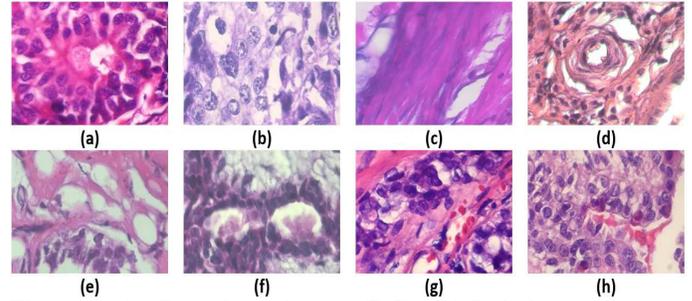}
\caption{(a) Ductal carcinoma (DC), (b) Lobular carcinoma (LC), (c) Mucinous carcinoma (MC), (d) Papillary carcinoma (PC), (e) Adenosis, (f) Fibroadenoma, (g) Tubular adenoma (TA), (h) Phyllodes tumor (PT). (a) to (d) \& (e) to (g) corresponds to malignant and benign class respectively.}
\end{figure}
%%%%%%%%%%%%%%%%%%%%%%%%%%%%%%%%%%%%%%%%%%%%%%%%
To resolve these challenges, a reliable \& more accurate practical method for BC multi-classification via DNN is developed. The proposed method discards feature engineering \& employs end-to-end training of deep discriminative ensemble of holistic CSML of HI \& local high-level to low-level semantic hierarchical hash signatures followed by softmax layer for classification. The model is validated with 7909 HI's to demonstrate its potency for deployment in clinical settings on generic processors \& experimental result shows superior performance. The main contributions can be recapitulated as:
\begin{itemize}
    \item A novel \& scalable DNN for BC multi-classification framework is explored. The model, as shown in Fig. 2, yields the highest recognition rate with reduced computational time \& complexity as compared to state-of-art. 
    \item The scheme strengthen the intra-class morphological similarities \& inter-class dissimilarities of the hierarchical feature space in a discriminative manner via deep ensemble of CSML \& heterogeneous local hash signatures.
\end{itemize}
%%%%%%%%%%%%%%%%%%%%%%%%%%%%%%%%%%%%%%%%%
\begin{figure*}[!htp]
\centering
\includegraphics[width=18cm, height=4.3cm]{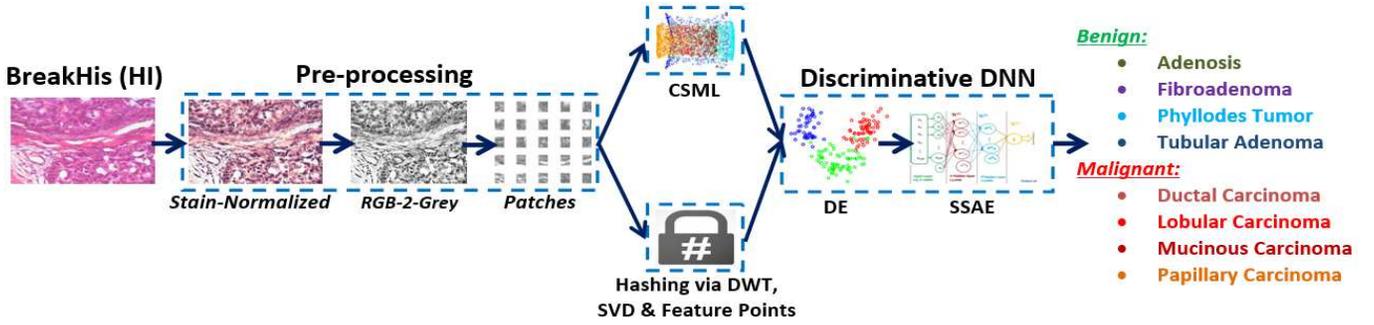}
\caption{Stages of integrated pipeline: pre-processing, CSML \& HI hash extraction, discriminative ensemble \& DNN learning.}
\end{figure*}
%%%%%%%%%%%%%%%%%%%%%%%%%%%%%%%%%%%%%%%%%%%%%%%%%%%%%
\section{Materials \& Prior Work}
\subsection{\textit{BreakHis} Dataset}
The proposed BC multi-classification method is examined on publicly available large-scale \textit{BreaKHis} dataset [8] containing 7,909 H\&E stained microscopic images from surgical biopsy (SOB) breast tumors, taken from 82 patients \& collected at multiple magnification factor of: 40$\times$,100$\times$,200$\times$ \& 400$\times$. The images are of 700$\times$460 pixels dimension with 24-bit color depth in 3-channel RGB (Red-Green-Blue) format. Table 1 summarizes the distribution of different histological sub-types \& a detailed description can be traced from [8]. %%%%%%%%%%%%%%%%%%%%%%%%%%%%%%%%%%%%%%%%%%%%%%%%
\begin{table}[!ht]
  \begin{center}
    \caption{Distribution of classes, sub-classes in \textit{BreaKHis} @ different magnification factors.}
    \begin{tabular}{|c|c|c|c|c|c|c|c|}
    \toprule
     &&\multicolumn{4}{ |c |}{\textbf{Magnification Factor}}&&\\
    \cmidrule{3-6}
    \textbf{Class} &\textbf{SC} & \textbf{40$\times$} & \textbf{100$\times$} & \textbf{200$\times$} & \textbf{400$\times$} &\textbf{Total}&\textbf{Patient}\\
    \midrule
    &DC&864&903&896&788&\textbf{3451}&\\
    \cmidrule{2-7}
    \textbf{M}&LC&156&170&163&137&\textbf{626}&\\
    \cmidrule{2-7}
    &MC&205&222&196&169&\textbf{792}&58\\
    \cmidrule{2-7}
    &PC&145&142&135&138&\textbf{560}&\\
    \midrule
    &A&114&113&111&106&\textbf{444}&\\
    \cmidrule{2-7}
    \textbf{B}&F&253&260&264&237&\textbf{1014}&\\
    \cmidrule{2-7}
    &TA&109&121&108&115&\textbf{453}&24\\
    \cmidrule{2-7}
    &PT&149&150&140&130&\textbf{569}&\\
    \midrule
     \multicolumn{2}{ |c |}{\textbf{Total}}&\textbf{1995}&\textbf{2081}&\textbf{2013}&\textbf{1820}&\textbf{7909}&\textbf{82}\\
    \midrule
    \end{tabular}
    \end{center}
\vspace{-8mm}
\end{table}
%%%%%%%%%%%%%%%%%%%%%%%%%%%%%%%%%%%%%%%%%%%%%%
\subsection{Prior Art using \textit{BreaKHis}}
Significant research is concentrated around binary classification of benign \& malignant classes. Over the past few years, researches have broadly investigated optimal feature engineering \& deep learning based architectures. These can be found from the works of Spanhol et al., [8], ensemble classifier (EC) of shallow features by Gupta et al., [9], multiple feature vector (MFV) \& transfer learning [10], graph-manifold \& BI-LSTM models by Pratiher et al., [11], Grassmann manifold (GM) aided vector of locally aggregated descriptors (VLAD) by Dimitropoulos et al., [12] and convolution neural network (CNN) model with fusion rule (FR) [13]. Efficacy of ConvNet based fisher vector (CFV) \& Gaussian mixture model (GMM) by Song et al., [14], deep CNN by Wei et al., [15] and BC classification using incremental boosting convolution networks in [16]. An exhaustive comparative study can be traced from Table III. Studies concerning multi-classification of sub-classes for clinical diagnosis or prognosis is done by Han Zhongyi et al., using class structured deep CNN (CSD-CNN) model [15] \& Bardou et al. using CNN based approach [17]. Multi-classification BC histological research comparison can be found in Table IV \& V.
%%%%%%%%%%%%%%%%%%%%%%%%%%%%%%%%%%%%%%%%%%%%%%%%%%%
\begin{figure*}[!htp]
\centering
\includegraphics[width=18cm, height=3.8cm]{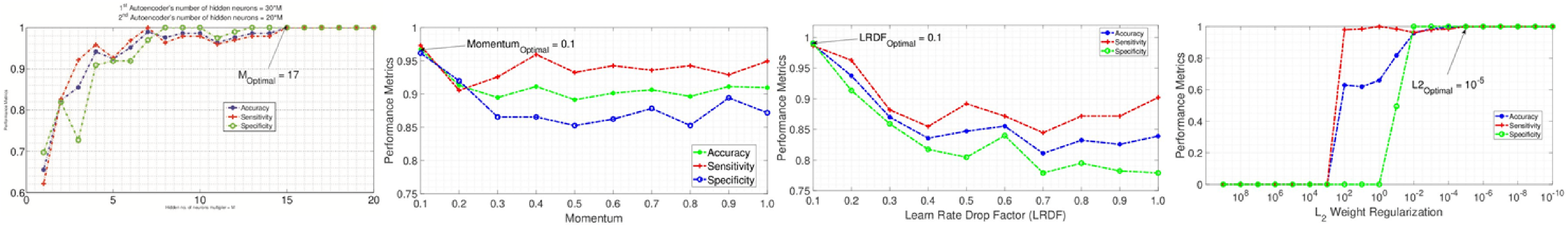}
\caption{Hyper-parameter tuning in DNN. (a) hidden size, (b) momentum, (c) learn rate drop factor, (d) $L_{2}$ weight regularization}
\end{figure*}
%%%%%%%%%%%%%%%%%%%%%%%%%%%%%%%%%%%%%%%%%%%%%%%%%%%
\section{Methodology \& Related Theory}
\subsection{Experiment Design: \textbf{Deep Discriminative Ensemble of Histological Hashing \& Class-Specific Manifold Learning}}
Recently, DNN has shown efficacy in achieving state-of-the-art performance in diverse research problems spanning medical imaging [4], natural language processing (NLP) \& speech processing. Here, we propose to use stacked sparse autoencoder (SSAE) based DNN [21] for robust BC multi-classification. Fig. 2 highlights the workflow. Our model pipeline consists of three main stages: Pre-processing stage for stain normalization, \& overlapping \& optimal patch segments of a specific size are generated for subsequent tissue index profile abstraction. Thereafter, class-specific manifold learning (CSML) of different histological sub-types are comprehended for nonlinear dimensionality reduction (NLDR). NLDR distillates discriminative low-dimensional structures pertinent to particular sub-class hidden in the high-dimensional HI. CSML preserves the intrinsic quasi-isometric geometry \& local contour connectivity of HI point-cloud within tolerable limits via feature-space geometry constraints \& is very much crucial for diagnosis. Thereafter, different hash signature obtained via discrete wavelet transform (DWT), singular-value decomposition (SVD) \& perceptual feature-points augments the local shallow statistical HI descriptors. The holistic CSML \& Hash vectors are fused in a discriminative fashion, which contemplates class structure based feature fusion. These ensemble discriminative super feature vectors are fed to SSAE for learning deep features \& classification thereof.
%%%%%%%%%%%%%%%%%%%%%%%%%%%%%%%%%%%%%%%%%%%%
\subsection{Class-Specific Manifold Learning (CSML)}
CSML is envisaged via Landmark Isomap (L-ISOMAP) aide Eigen sub-space estimation of a particular histological sub-type. For a vectorized HI point-cloud of $Y$ data-points, arbitrarily $m(m\ll M)$ points are selected as Landmarks. Euclidean distance based embedding of the input data using $m$ landmark points are selected from $Y$ \& followed by multi-dimensional scaling (MDS) using the $m\times m$ matrix $G_{m,M}$ of geodesic distances for each landmark pair to compute the low dimensional feature space. Mathematically, it is given by:
\begin{equation}
    m^T(p,q)=-\frac{1}{2} \left ( F^2_{pq}-e_p\frac{1}{m}\sum_lH^2_{pl} \right ),
\end{equation}
where, $H^2$ is the means geodesic distance matrix $H$ is element-wise squared \& $e_p$ is the Eigen vector with zero Eigen value. Details about L-ISOMAP can be traced from [22].
%\begin{equation}
 %   X^d_{1:m}=\sqrt{\frac{m}{n}}\begin{bmatrix}
%D\\m
%\end{bmatrix}V_{1:d}\sum^{-1}_{1:d}.
%\end{equation}
%%%%%%%%%%%%%%%%%%%%%%%%%%%%%%%%%%%%%%%%%%
\subsection{Histological Hashing for Local Signatures}
Rudimentary image hashing details can be found in [23-26]. Histological image hashing encodes locality reference \& inherent neighborhood connectivity of the underlying HI. Here, we have used the following hash signatures:
\subsubsection{Discrete Wavelet Transform (DWT) Based Image Hash} Computes robust \& compact hash via 2D DWT on HI, which decomposes into four sub-bands. Both edge or high frequency information \& coarse stable low-frequent coefficients are perceived via DWT coefficients [23].  
\subsubsection{Hashing Via Singular Value Decomposition (SVD)}
SVD based image hasing has robust tolerance to small rotational changes until 10$^\cdot$ \& is translation in-variance \& incorporates low rank approximation of original normalized sub-image of HI \& non-correlated directional feature space encoding [24].
\subsubsection{Feature Point Based Image Hashing} 
Statistical image features like Hessian affine, maximally stable extremal region (MSER) detectors, Harris corner detector \& feature points based on end-stopping behavior like end-stopped Wavelets such as Morlets preserves the momentous image geometry feature-space constraints of 2-D HI pixels \& mapped to 1-D feature vector which is compressed to generate the hash vector [25].
%%%%%%%%%%%%%%%%%%%%%%%%%%%%%%%%%%%%%%%%%%%%%%
\subsection{SSAE aided DNN Training \& Hyper-parameters Tuning}
Discriminative feature ensemble of holistic CSML \& shallow hash signatures is fused via discrimination correlation analysis [27]. Stacked Sparse Autoencoder (SSAE) [28] involves optimal parameter $\theta=(V, a_k, a_y)$ computation by minimizing the error between the model input and output. Rudimentary details can be traced from [21]. The model is trained on a generic system woth mode configuration as 3.50 GHz processor, 16GB RAM \& AMD FX-8320 Octa-core. Fig 3 shows hyper-tuned optimal SSAE parameters. A maximum epoch of 500 \& number hidden layer neurons in the first and second autoencoders (AE) is kept at 500 \& 300 respectively. Further, $L_{2}$ weight regularization, sparsity regularization \& sparsity proportion in the two AE are set to 0.001, 4 \& 0.15. '\textit{tanh}' activation function (AF) in the hidden units is connected to by fully connected layer \& classification done via a softmax layer. '\textit{tanh}' AF is used due to its robust tolerance to approximate intrinsic manifold non-linearity \& extraction of mutual dependence for further segregation thereof. Piece-wise back propagation learning via stochastic gradient descent (SGD) algorithm is envisaged with a learning rate of $10^{-4}$ \& the initial random weights drawn uniformly from $\left [-0:1; 0:1 \right]$, gradient decay factor of 0.2, momentum of $0.6$, learning rate drop period (LRDP) of 5 \& $L_{2}$ weight regularization of $10^{-4}$ is used during the training phase. The weights (W) \& bias (b) are updated as
\begin{equation}
    W_{l}=W_{l}-\eta\frac{\partial}{\vartheta W_l}{(W,b;X,t)}=B_lW_l-\eta\frac{\partial}{\vartheta B_l}{(W,B;X,t)}
\end{equation}
%    W_{l}=W_{l}-\eta\frac{\partial}{\vartheta W_l}{(W,b;X,t)}\\=B_lW_l-\eta\frac{\partial}{\vartheta B_l}{(W,B;X,t)}
where, $W_l$  represents the weights, $B_l$  represents the $l^{th}$ layer bias, $\eta$ is learning rate, (X,t) is the mini-batch comprising of \textit{'m'} training samples. In order to eschew skewed model over-fitting \& bias minimization, the data is randomly partitioned in the ratio 60:20:20 for to training, validation \& testing phase. 
%%%%%%%%%%%%%%%%%%%%%%%%%%%%%%%%
\section{Results and Discussion}
The adequacy of the proposed method is evaluated in terms of standard evaluation metrics i.e., classification accuracy (AC), sensitivity (SN) \& specificity (SP). Experimental results for both binary \& multi-classification are compared with state-of-the-art techniques on benchmark \textit{BreaKHis} dataset. 
%%%%%%%%%%%%%%%%%%%%%%%%%%%%%%%%%%%%%%%%%%%%%%%%%
\begin{table}[!ht]
\centering
\caption{Results for Binary Classification. PM = Performance Metrics}
\begin{tabular}{@{}|c|c|c|c|c|c|@{}}
\midrule
\begin{tabular}[x]{@{}c@{}}\textbf{Class}\end{tabular} & \begin{tabular}[x]{@{}c@{}}\textbf{PM} \\ \textbf{(\%)} \end{tabular} &\multicolumn{4}{ c|}{\textbf{Magnification Factor}}\\ 
\cmidrule{3-6}
\textbf{}& \textbf{} &\textbf{40X} & \textbf{100X} & \textbf{200X} & \textbf{400X}\\
\midrule
\multirow{4}{*}{Binary} & AC & 99.1 & 98.7& \textbf{99.3} & 98.4\\ \cmidrule(l){2-6} 
 & SN & \textbf{100} & \textbf{100} &\textbf{\textbf{100}} &\textbf{100}\\ \cmidrule(l){2-6} 
 & SP & 98.3 & 97.9 & \textbf{98.5} & 96.8 \\ \midrule
 \end{tabular}
\vspace{-5mm}
\end{table}
%%%%%%%%%%%%%%%%%%%%%%%%%%%%%%%%%%%%%%%%%%%%%%%%%%%%%%%%%%%%%%%
\subsection{Grading Tumor Malignancy: Binary Classification}
Initially, binary classification of BC into benign \& malignant classes is done to ensure its competence in coarse HI characterization. Table II highlights the performance measure of the experimental results for different magnification factors. 100\% sensitivity for all magnification factor ensures that all malignant classes are recognized correctly. As such, pathologist may invest more time for identifying benign cases with our system \& demonstrates the effectiveness of the DNN model to learn deep malignancy biomarkers for discriminative HI grading. A comparative study of the proposed mode with the existing state-of-the-art methods [8-20] is given in Table III. It may be noted that the proposed framework outperforms all the previously used methods in terms of classification accuracy with a significant enhancement at all magnification factors. Further, our proposed discriminative DNN framework outperforms the conventional \& visual feature descriptor based approach such LBP [8], VLAD [12] and KAZE [20] \& also surpasses recent CNN \& BiLSTM based methods [10-11][14-16], which is proved to be optimal in analyzing visual imagery. This indicates that our proposed system which is robust in terms of performance \& computationally much more efficient as it runs on generic laptops as compared to CNN models which requires graphics processing unit (GPU).
%%%%%%%%%%%%%%%%%%%%%%%%%%%%%%%%%%%%%%%%%%%%%%%
\begin{table}[!ht]
\centering
\caption{State-of-the-art comparison for binary classification of benign and malignant classes}
\begin{tabular}{@{}|l|l|c|c|c|c|@{}}
\midrule
\begin{tabular}[x]{@{}c@{}}\textbf{Ref,}\\\textbf{Years} \end{tabular} & \begin{tabular}[x]{@{}c@{}}\textbf{Feature + Method} \end{tabular} &\multicolumn{4}{ c |}{\textbf{Performance (\%)}}\\ 
\cmidrule{3-6}
\textbf{}& \textbf{} & \textbf{40X} & \textbf{100X} & \textbf{200X} & \textbf{400X}\\
\midrule
$[13]$, 2016&\begin{tabular}{@{}c@{}} CNN, FR \end{tabular}&85.6&83.5&82.7&80.7\\
\midrule
$[8]$ 2016,&\begin{tabular}{@{}c@{}} CLBP, SVM \end{tabular}&77.40&76.40&70.20&72.80\\
\midrule
$[9]$, 2017&\begin{tabular}{@{}c@{}} C-TID, EC \end{tabular}&87.2&88.22&88.89&85.82\\
\midrule
$[10]$, 2017&\begin{tabular}{@{}l@{}} CNN, DeCAF, \\MFV \end{tabular}&84.6&84.8&84.2&81.6\\
\midrule
$[11]$, 2017&\begin{tabular}{@{}c@{}} GML, BI-LSTM\end{tabular}&96.2&97.2&97.1&95.4\\
\midrule
$[12]$, 2017&\begin{tabular}{@{}c@{}} VLAD, GM \end{tabular}&91.8&92.1&91.4&90.2\\
\midrule
$[14]$, 2017&\begin{tabular}{@{}l@{}} I-EM, CFV, CNN, \\GMM \end{tabular}&87.7&87.6&86.5&83.9\\
\midrule
$[15]$, 2017&\begin{tabular}{@{}c@{}} CSDCNN \end{tabular}&95.8&96.9&96.7&94.9\\
\midrule
$[16]$, 2018&\begin{tabular}[x]{@{}l@{}} DCNN \end{tabular}&95.1&96.3&96.9&93.8\\
\midrule
$[17]$, 2018&\begin{tabular}[x]{@{}l@{}}Dense SIFT, SURF, \\BOW, LCLC \end{tabular}&98.33&97.12&97.85&96.15\\
\midrule
$[18]$, 2018&\begin{tabular}{@{}l@{}} FV, CSE\end{tabular}&87.5&88.6&85.5&85.0\\
\midrule
&\begin{tabular}{@{}l@{}} TL, DenseNet\end{tabular}&84.72&89.44&95.65&82.65\\
\cmidrule{2-6}
$[19]$, 2018&\begin{tabular}{@{}l@{}} CNN, DenseNet \end{tabular}&91.90&93.64&95.84&90.15\\
\cmidrule{2-6}
&\begin{tabular}{@{}l@{}} MV, XGboost\end{tabular}&94.71&95.9&96.76&89.11\\
\midrule
$[20]$&\begin{tabular}{@{}l@{}} KAZE, BOF, \\binary SVM \end{tabular}&85.9&80.4&78.1&71.1\\
\midrule
\textbf{This Work}&\begin{tabular}{@{}l@{}} \textbf{CSML, Hashing}\\ \textbf{DNN}\end{tabular}&\textbf{99.1} & \textbf{98.7}& \textbf{99.3} & \textbf{98.4}\\
\midrule
\end{tabular}
\vspace{-5mm}
\end{table}
%%%%%%%%%%%%%%%%%%%%%
\subsection{Multi-classification Performance}
It's evident from literature survey that there are very few literature available in the multi-class classification of BC from HI \& as such, we further examine the proposed framework towards the multi-class classification of BC to demonstrate its efficacy towards practical use from a clinical perspective. State-of-the-art comparative evaluation is given in Table IV, whereas Table V gives class specific performance accuracy. It is evident that the proposed discriminative DNN framework not only surpasses the state-of-art on the basis of magnification factor but also on the basis of HI sub-classes in terms of recognition rate. Earlier in the state-of-art, CSDCNN [15] based approach exhibited best recognition rate for multi-class classification but our proposed method surpasses it in all the magnification factor by \textbf{3.3\%, 1.8\%, 2.6\%, 3.5\%}, for $40\times$, $100\times$, $200\times$ \& $400\times$ magnification. Table V shows that our method surpasses the class-specific recognition rate with high margin \& in some case as high as \textbf{38.2\%} enhancement for Lobular carcinoma (LC) sub-class.
%%%%%%%%%%%%%%%%%%%%%%%%%%%%%%%%%%%%
\begin{table}[!ht]
\centering
\caption{State-of-the-art comparison for multi-Classification histological sub-types}
\begin{tabular}{@{}|l|l|c|c|c|c|@{}}
\midrule
\begin{tabular}[x]{@{}c@{}}\textbf{Ref}\\\end{tabular} & \begin{tabular}[x]{@{}c@{}}\textbf{Feature + Method} \end{tabular} &\multicolumn{4}{ c |}{\textbf{Performance (\%)}}\\ 
\cmidrule{3-6}
\textbf{}& \textbf{} & \textbf{40X} & \textbf{100X} & \textbf{200X} & \textbf{400X}\\
\midrule
$[15]$&\begin{tabular}{@{}c@{}} CSDCNN \end{tabular}&92.8&93.9&93.7&92.9\\
\midrule
&\begin{tabular}{@{}c@{}} DSIFT $+$ BoW \end{tabular}&41.80&38.56&49.75&38.67\\
\cmidrule{2-6}
&\begin{tabular}{@{}c@{}} SURF $+$ BoW \end{tabular}&53.07&60.80&70.00&51.01\\
\cmidrule{2-6}
&\begin{tabular}{@{}c@{}} DSIFT $+$ LLC \end{tabular}&60.58&57.44&70.00&49.96\\
\cmidrule{2-6}
&\begin{tabular}{@{}c@{}} SURF $+$ LLC \end{tabular}&80.37&63.84&74.54&54.70`\\
\cmidrule{2-6}
\begin{tabular}{@{}l@{}} $[17]$ \end{tabular}&\begin{tabular}{@{}c@{}} DSIFT, BoW $+$SVM\end{tabular}&18.77&17.28&20.16&17.49\\
\cmidrule{2-6}
&\begin{tabular}{@{}c@{}} SURF, BoW$+$ SVM \end{tabular}&49.65&47.00&38.84&29.50\\
\cmidrule{2-6}
&\begin{tabular}{@{}c@{}} DSIFT, LLC$+$SVM\end{tabular}&48.46&49.44&43.97&32.60\\
\cmidrule{2-6}
&\begin{tabular}{@{}c@{}} SURF, LLC$+$ SVM \end{tabular}&55.80&54.24&40.83&37.20\\
\cmidrule{2-6}
&\begin{tabular}{@{}c@{}} CNN, SVM-RBF \end{tabular}&75.43&71.20&67.27&65.12\\
\midrule
\begin{tabular}{@{}c@{}} \textbf{This}\\\textbf{Work} \end{tabular}&\begin{tabular}{@{}c@{}} \textbf{CSML, Hashing, DNN} \end{tabular}&\textbf{95.1}&\textbf{95.7}&\textbf{95.8}&\textbf{95.2}\\
\midrule
\end{tabular}
\end{table}
%%%%%%%%%%%%%%%%%%%%%%%%%%%%%%%%%%%%%%%%%%%%%%%
\begin{table}[!ht]
  \begin{center}
    \caption{Class-specific performance comparison with [17]}
    \begin{tabular}{|c|c|c|c|c|c|c|}
    \toprule
     &&&\multicolumn{4}{ |c |}{\textbf{Magnification Factor}}\\
    \cmidrule{4-7}
    \textbf{Class}&\textbf{Ref, years} &\textbf{SC} & \textbf{40$\times$} & \textbf{100$\times$} & \textbf{200$\times$} & \textbf{400$\times$}\\
    \midrule
    &\begin{tabular}[x]{@{}r@{}}$[17]$, 2018 \end{tabular}&DC&91.51&90.77&91.14&92.74\\
    \cmidrule{2-7}    
    &\textbf{This Work}&DC&\textbf{96.7}&\textbf{97}&\textbf{97.6}&\textbf{96.9}\\
    \cmidrule{2-7}
    Malignant&\begin{tabular}[x]{@{}r@{}}$[17]$, 2018 \end{tabular}&LC&78.72&54.90&63.27&56.10\\
    \cmidrule{2-7}    
    &\textbf{This Work}&LC&\textbf{93.8}&\textbf{94.7}&\textbf{92.8}&\textbf{93.1}\\
    \cmidrule{2-7}
    &\begin{tabular}[x]{@{}r@{}}$[17]$, 2018 \end{tabular}&MC&70.49&82.09&61.02&70.59\\
    \cmidrule{2-7}    
    &\textbf{This Work}&MC&\textbf{94.4}&\textbf{95.8}&\textbf{96.6}&\textbf{94.9}\\
    \cmidrule{2-7}
    &\begin{tabular}[x]{@{}r@{}}$[17]$, 2018 \end{tabular}&PC&67.44&83.72&57.50&68.29\\
    \cmidrule{2-7}    
    &\textbf{This Work}&PC&\textbf{93.1}&\textbf{95.2}&\textbf{93}&\textbf{93.1}\\
\midrule
    &\begin{tabular}[x]{@{}r@{}}$[17]$, 2018 \end{tabular}&A&85.29&79.41&84.85&90.63\\
    \cmidrule{2-7}    
    &\textbf{This Work}&A&\textbf{93.5}&\textbf{93.9}&\textbf{97}&\textbf{94.1}\\
    \cmidrule{2-7}
    Benign&\begin{tabular}[x]{@{}r@{}}$[17]$, 2018 \end{tabular}&F&86.84&91.03&91.14&77.46\\
    \cmidrule{2-7}    
    &\textbf{This Work}&F&\textbf{95}&\textbf{95.6}&\textbf{94.9}&\textbf{95.1}\\
    \cmidrule{2-7}
    &\begin{tabular}[x]{@{}r@{}}$[17]$, 2018 \end{tabular}&TA&75.56&93.33&76.19&82.05\\
    \cmidrule{2-7}
    &\textbf{This Work}&TA&\textbf{94.5}&\textbf{94.1}&\textbf{94.3}&\textbf{94.4}\\
    \cmidrule{2-7}
    &\begin{tabular}[x]{@{}r@{}}$[17]$, 2018 \end{tabular}&PT&76.19&63.89&62.50&58.82\\
    \cmidrule{2-7}
    &\textbf{This Work}&PT&\textbf{94.4}&\textbf{95.1}&\textbf{95.2}&\textbf{94.7}\\
    \midrule
    \end{tabular}
    \vspace{-8mm}
    \end{center}
\end{table}
%%%%%%%%%%%%%%%%%%%%%%%%%%%%%%%%%%%%%%%%%%%%%%%%
\section{CONCLUSION}
A novel deep discriminative ensemble learning CAD for multi-class BC characterization is introduced in this work. The method implements deep contextual grading of hybrid holistic-level CSML representations \& local hash signatures of HI, thereby, effectively discriminating between benign \& malignant sub-classes. The proposed approach was validated using \textit{BreakHis} dataset \& experimental results exemplify superior discriminating performance as compared to the existing state-of-the-art. In particular, it shows high specificity towards malignant sub-classes, which can assist pathologists by reducing their heavy workload \& arrange optimal therapeutic schedules for further diagnosis or prognosis of benign tissues.   

Currently, the method is being escalated to include deeper structures using graph CNN \& sequential contextual learning with other tissue images to investigate the diagnostic modality.
%%%%%%%%%%%%%%%%%%%%%%%%%%%%%%%%%%%%%%%
\addtolength{\textheight}{-7.5cm}  

\end{document}